\documentclass[11pt,a4paper]{article}
\usepackage[hyperref]{eacl2021}
\usepackage{times}
\usepackage{latexsym}

\usepackage{microtype}
\usepackage{amsmath}
\usepackage{amsthm}
\usepackage{amssymb}
\usepackage{booktabs}
\usepackage{multirow}
\usepackage{graphicx}
\usepackage{caption}
\usepackage{subcaption}
\usepackage{makecell}
\aclfinalcopy 
\def\aclpaperid{1045} 
\newcommand{\vect}[1]{\mathbf{#1}}

\newcommand{\vx}[0]{\vect{x}}
\newcommand{\vy}[0]{\vect{y}}

\DeclareMathOperator*{\argmax}{arg\,max}

\title{On Hallucination and Predictive Uncertainty in\\ Conditional Language Generation}

\author{Yijun Xiao, William Yang Wang\\
University of California, Santa Barbara\\
  {\tt \{yijunxiao,william\}@cs.ucsb.edu} \\}

\date{}

\begin{document}
\maketitle
\begin{abstract}
Despite improvements in performances on different natural language generation tasks, deep neural models are prone to hallucinating facts that are incorrect or nonexistent. Different hypotheses are proposed and examined separately for different tasks, but no systematic explanations
are available across these tasks. In this study, we draw connections between hallucinations and predictive uncertainty in conditional language generation. We investigate their relationship in both image captioning and data-to-text generation and propose a simple extension to beam search to reduce hallucination. Our analysis shows that higher predictive uncertainty corresponds to a higher chance of hallucination. Epistemic uncertainty is more indicative of hallucination than aleatoric or total uncertainties. It helps to achieve better results of trading performance in standard metric for less hallucination with the proposed beam search variant.
\end{abstract}

\section{Introduction}
Modern deep neural network models have brought drastic improvements of generation quality measured by standard metrics on different natural language generation (NLG) tasks. However, along with these improvements, researchers find that neural models are more prone to a phenomenon called hallucination, where models generate description tokens that are not supported by the source inputs. This phenomenon seriously damages the applicability of neural language generation models in practice where information accuracy is vital.

Hallucination has been observed in various conditional NLG tasks such as image captioning \cite{rohrbach2018object}, data-to-text generation \cite{wiseman2017challenges,nie2019simple,parikh2020totto}, abstractive summarization \cite{cao2018faithful,durmus2020feqa}, and neural machine translation (NMT) \cite{muller2019domain}. These studies tackle hallucinations within a specific task and give possible explanations of why hallucinations occur. For example, \citet{rohrbach2018object} attributes object hallucination in image captioning to visual misclassification and over-reliance on language priors; \citet{nie2019simple} believes hallucination in neural surface realization comes from the misalignment between meaning representations and their corresponding references in the dataset; \citet{muller2019domain} claims that hallucinations in NMT are mainly due to domain shift.

We believe that there is a common theme across all the hallucination explanations in conditional NLG tasks: predictive uncertainty.
In language generation, predictive uncertainty quantifies the entropy of the token probability distributions a model predicts. There are multiple sources of uncertainty. Two major ones frequently studied are aleatoric and epistemic uncertainties, where the former comes from the data or measurements, and the latter is concerned with the model. With recent progress in Bayesian neural networks (BNNs) \cite{hinton1993keeping,neal1995bayesian} and uncertainty quantification \cite{blundell2015weight,gal2016dropout,lakshminarayanan2017simple}, we are able to quantify both parts of predictive uncertainty in neural NLG.

This study draws connections between hallucination and predictive uncertainty and empirically investigates their relationship in image captioning and data-to-text generation tasks. We propose an uncertainty-aware beam search algorithm to reduce the chance of hallucination by penalizing parts or the entirety of the predictive uncertainty during model decoding. We find that the choice of uncertainty matters, and penalizing epistemic uncertainty yields better results compared to penalizing aleatoric or total uncertainty. Our contributions are:
\begin{itemize}
    \item We draw connections between hallucination and predictive uncertainty across various conditional natural language generation tasks and empirically investigate their relationship.
    \item We propose an uncertainty-aware beam search approach for hallucination reduction to demonstrate that lowering uncertainty can lead to less hallucination.
    \item We show that uncertainty decomposition helps to achieve better trade-offs between hallucination and performance.
\end{itemize}

\section{Hallucination and Predictive Uncertainty}

\subsection{Hallucination Probability}
In general, hallucination refers to the phenomenon where the model generates false information not supported by the input. For example, in the context of image captioning, hallucination can be defined as generating captions that contain descriptions not present in the given image. 
Let $(x,y)$ be the pair of variables at interest where $x$ is some structured data containing facts and $y$ is a natural language sentence based on the facts. The task is to learn the conditional distribution of $p(y|x)$ in order to generate sentence $y$ given any new input $x$. Most neural approaches break the probability into a sequence of single token predictions:
\begin{align}
    p(y|x)=p(y_1|x)\prod_{i=2}^kp(y_i|x,y_1,\cdots,y_{i-1})
\end{align}
where $\{y_1,\cdots,y_k\}$ is the collection of tokens in sentence $y$. We denote $c_i=\{x,y_1,\cdots,y_{i-1}\}$ as the context of the $i$-th prediction in the following sections for simplicity.

Apparently, hallucination is context-dependent which means we need to look at a certain context $c_i$ and determine whether the next token prediction $y_i$ is hallucinated or not. Let $\mathcal{V}_h^{(c_i)}$ denote the set of tokens that are considered false information given the current context $c_i$ and $\mathcal{V}$ the whole vocabulary. Consider a random sampling decoder where a token is generated based on the predicted categorical distribution. i.e. $\text{Cat}(|\mathcal{V}|,p(y_i|c_i))$. The probability of hallucination at the current step is simply:
\begin{align}
    P(y_i\in\mathcal{V}_h^{(c_i)}) = \sum_{v\in\mathcal{V}_h^{(c_i)}}p(y_i=v|c_i)
\end{align}

Practically, it is hard to automatically determine the context-dependent set $\mathcal{V}_h^{(c_i)}$. Task-specific heuristics are often used to determine which tokens are hallucinated. In specific restrictive applications, the context-dependent set can be relaxed to a context-independent one to reduce the complexity of determining hallucination.

\subsection{Relationship with Predictive Uncertainty}
\label{sec:pred_unc}
We use entropy to measure the predictive uncertainty in this work. The total uncertainty of predicting token $y_i$ is:
\begin{align}
\label{equ:entropy}
    &H(y_i|c_i) \nonumber\\
    =&-\sum_{v\in\mathcal{V}}p(y_i=v|c_i)\log{p(y_i=v|c_i)} \nonumber\\
    =&-\sum_{v\in\mathcal{V}\setminus\mathcal{V}_h^{(c_i)}}p(y_i=v|c_i)\log{p(y_i=v|c_i)} \nonumber\\
    &-\sum_{v\in\mathcal{V}_h^{(c_i)}}p(y_i=v|c_i)\log{p(y_i=v|c_i)}
\end{align}

From Equation \ref{equ:entropy}, we can see that there are two sources of uncertainty for the token predictions: one from the uncertainty of choosing suitable tokens to describe the input; another from some unsuitable tokens attaining considerable probability mass either by being confusing in the current context or due to an insufficiently trained system.

The second source of uncertainty is directly related to hallucination probability. Although no monotonic relationship can be derived, a near-zero hallucination probability requires a near-zero value of the second source of uncertainty. This observation prompts us to investigate the relationship between hallucination and predictive uncertainty in practice. Intuitively, the higher the predictive uncertainty is, the more probable some of the probability mass gets assigned to unsuitable tokens.

\begin{figure}[t]
    \centering
    \begin{subfigure}[t]{0.48\textwidth}
        \centering
        \includegraphics[width=\textwidth]{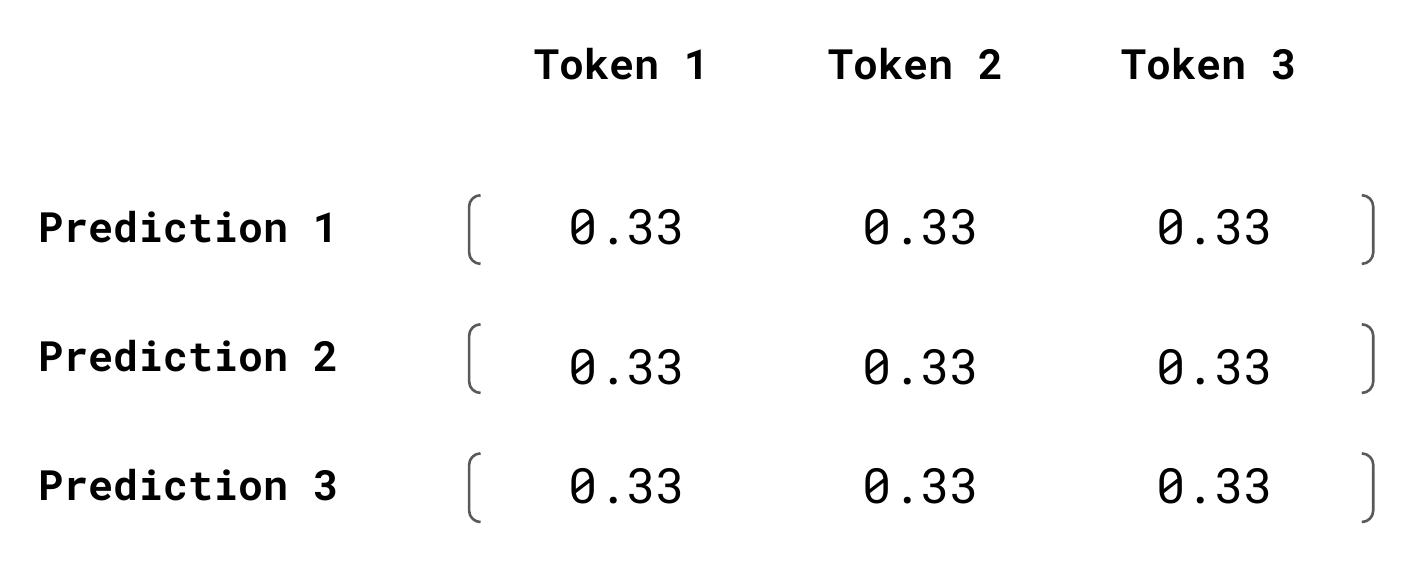}
        \caption{}
    \end{subfigure}
    \vfill
    \begin{subfigure}[t]{0.48\textwidth}
        \centering
        \includegraphics[width=\textwidth]{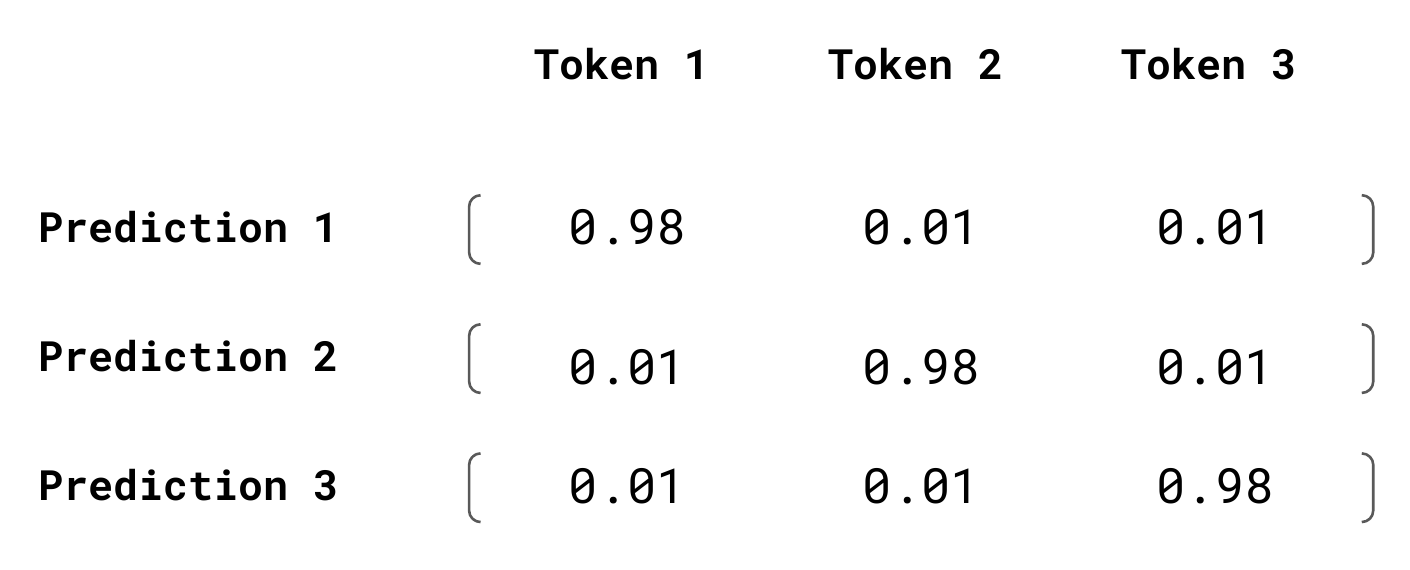}
        \caption{}
    \end{subfigure}
    
    \caption{Examples of predictions with (a) high aleatoric but low epistemic uncertainty; and (b) high epistemic but low aleatoric uncertainty.}
    \label{fig:ep_al}
\end{figure}

\subsection{Uncertainty Decomposition}
There are often two types of uncertainties frequently mentioned in uncertainty quantification literature: epistemic and aleatoric uncertainty \cite{der2009aleatory,kendall2017uncertainties,depeweg2018decomposition}. Epistemic uncertainty reflects the uncertainty on model weights, and aleatoric uncertainty concerns inherent uncertainty in the data or measurement. We are interested in whether the relationship with hallucination is the same for both types of uncertainties.

Bayesian deep learning approaches \cite{blundell2015weight,gal2016dropout,lakshminarayanan2017simple} are widely studied for uncertainty quantification with neural networks. Following the notations in Section \ref{sec:pred_unc}, the predictive distribution of $p(y_i|c_i)$ can be written as:
\begin{align}
    p(y_i|c_i)=\int_w p(y_i|c_i,w)q(w)dw
\end{align}
where $w$ parameterizes the neural network that makes predictions and $q(w)$ denotes the approximate posterior distribution of the weights $w$ given the training data. Notice that if we fix the weights $w$, $H(y_i|c_i,w)$ represents the entropy that is unrelated to the uncertainty of the model weights. Therefore the aleatoric part of the predictive uncertainty can be calculated with $\mathbb{E}_{q(w)}[H(y_i|c_i,w)]$. The epistemic part of the uncertainty is the difference between the total and the aleatoric uncertainty as shown below:
\begin{align}
    u_{al}(y_i|c_i)&=\mathbb{E}_{q(w)}[H(y_i|c_i,w)] \\
    u_{ep}(y_i|c_i)&=H(y_i|c_i)-\mathbb{E}_{q(w)}[H(y_i|c_i,w)]
\end{align}

In this study, the aleatoric and epistemic parts of predictive uncertainty are estimated using deep ensembles \cite{lakshminarayanan2017simple}. More concretely, denote the model predictions as $\{p_m(y_i|c_i)\}_{m=1}^M$ and the aggregated prediction as $p(y_i|c_i)=\frac{1}{M}\sum_{m=1}^Mp_m(y_i=v|c_i)$, aleatoric and epistemic uncertainties are calculated as:
\begin{align}
    u_{al}(y_i|c_i)&=\frac{1}{M}\sum_{m=1}^MH_m(y_i|c_i) \\
    u_{ep}(y_i|c_i)&=H(y_i|c_i)-u_{al}(y_i|c_i)
\end{align}
where $H_m(y_i|c_i)$ and $H(y_i|c_i)$ are the entropy of $p_m(y_i|c_i)$ and $p(y_i|c_i)$ respectively.

Intuitively, in the case of deep ensembles, aleatoric uncertainty measures the average spread of all model predictions, while epistemic uncertainty measures the agreement among all model predictions. Examples with three possible tokens are illustrated in Figure \ref{fig:ep_al}.

\begin{table*}[t]
    \small
    \centering
    \begin{tabular}{lcccccc}
    \toprule
         \multirow{2}{*}{\textbf{Model}} & \multicolumn{5}{c}{\textbf{Action hallucination \% at uncertainty level}} \\
         & $\leq 0.8$ & $0.8$ - $1.6$ & $1.6$ - $2.4$ & $2.4$ - $3.2$ & $3.2$ - $4.0$ & $> 4.0$ \\
         \midrule
         FC & 0.00 & 0.00 & 2.27 & 12.86 & 15.71 & 31.03 \\
         Att2In & 0.00 & 0.00 & 3.39 & 6.58 & 12.07 & 22.03 \\
         BUTD & 0.00 & 2.94 & 1.92 & 12.77 & 17.24 & 25.53 \\
         Transformer & 2.99 & 5.48 & 6.58 & 8.82 & 12.00 & 43.75 \\
         \bottomrule
    \end{tabular}
    \caption{Action hallucination percentages at different levels of predictive uncertainty. Action predictions with higher uncertainty are more prone to hallucination.}
    \label{tab:hal_vs_uncert}
\end{table*}

\section{Case Study: Image Captioning}
\label{sec:img_cap}
In this section, we analyze image captioning models trained on MSCOCO \cite{chen2015microsoft} data set. 

\subsection{Hallucination Probability at Different Uncertainty Levels}
The first question we want to investigate is whether hallucination probabilities change at different predictive uncertainty levels. Some experimental settings are listed below.

\paragraph{Model architecture} We consider four different image captioning models: \textbf{FC} model \cite{rennie2017self} where image features are used to initialize the RNN decoder; \textbf{Att2In} model from \cite{rennie2017self} applies attention on image features and feeds it into the decoder LSTM \cite{hochreiter1997long} cell gate; \textbf{BUTD} model from \cite{anderson2018bottom} uses bottom-up attention which operates at the level of objects and other salient image regions; \textbf{Transformer} model where transformers \cite{vaswani2017attention} are used in the encoder-decoder structure for generation. All models are implemented in the open source framework by \citet{luo2018discriminability}\footnote{https://github.com/ruotianluo/self-critical.pytorch}.

\paragraph{Training} We consider the same data split from \cite{karpathy2015deep}. All models are trained with batch size 50 for 30 epochs with Adam optimizer \cite{kingma2014adam}. Evaluations are done on the Karpathy test set.

\paragraph{Hallucination and uncertainty evaluation} As in \cite{rohrbach2018object}, synonyms for all possible MSCOCO objects are used to determine whether an object generated by the captioning model is hallucinated. Hallucination probabilities are calculated by binning all object token prediction entropy and counting the percentage of hallucinated objects in each bin.

\begin{figure}
    \centering
    \includegraphics[width=0.45\textwidth]{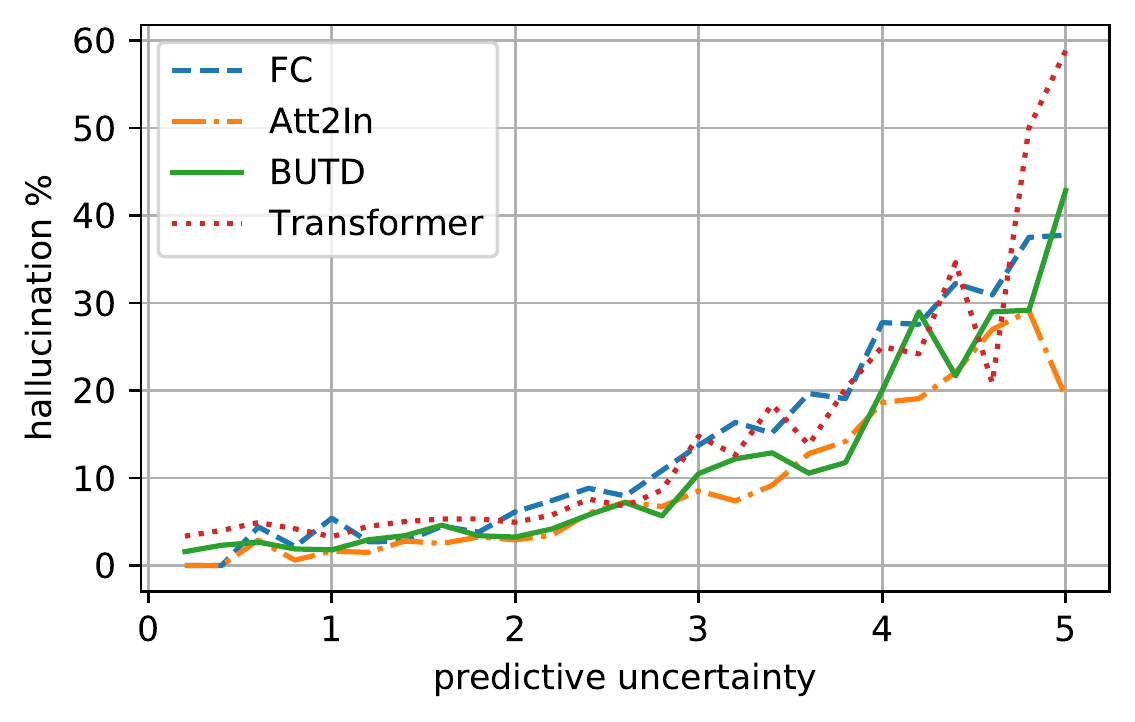}
    \caption{Object hallucination chance at different predictive uncertainty levels. Higher predictive uncertainty corresponds to a higher level of hallucination percentage across all models.}
    \label{fig:hal_vs_uncert}
\end{figure}

\subsection{Results and Discussions} 

Figure \ref{fig:hal_vs_uncert} shows the object hallucination percentages at different predictive uncertainty levels. At higher uncertainty levels, the generated objects are more likely to be hallucinated. The results are consistent across four different models. The transformer model seems to have a higher hallucination chance at high uncertainty levels than the other three models. 
However, this does not indicate Transformer models hallucinate more. In fact, the transformer model has an overall lowest hallucination percentage among all four models.

\paragraph{Beyond object hallucination}
Aside from object hallucination, we also analyze verbs generated by the models to see whether a similar relationship holds for other types of token generations. The same models and training procedures are adopted. We extract all present continuous tense verbs from the generated captions using spaCy part-of-speech tagger\footnote{https://spacy.io} and manually label whether they are suitable to describe the corresponding images. There are approximately 3500 generated captions containing verbs, and 400 are annotated for each model. We refer to unsuitable verbs generated in the captions as action hallucinations.

Action predictions are binned according to their uncertainty values, and the results are shown in Table \ref{tab:hal_vs_uncert}. We can observe that action tokens with higher predictive uncertainty are also more likely to be hallucinated. Noticeably, the transformer model also has a higher action hallucination rate at high uncertainty levels.

\begin{figure*}[t]
    \centering
    \begin{subfigure}[t]{0.2\textwidth}
        \centering
        \includegraphics[width=\textwidth]{figs/COCO_val2014_000000136572.jpg}
        \caption{a red and black \textbf{motorcycle} (\textit{0.58}) parked in a parking lot}
    \end{subfigure}
    \hfill
    \begin{subfigure}[t]{0.2\textwidth}
        \centering
        \includegraphics[width=\textwidth]{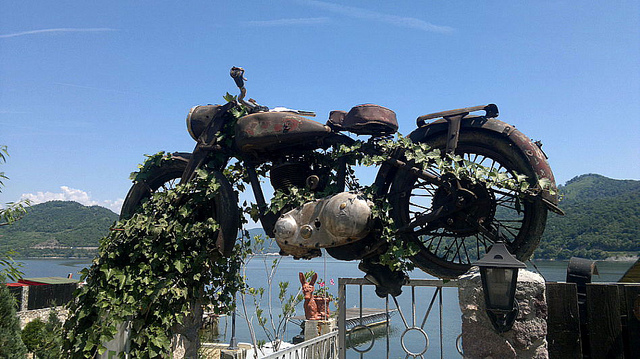}
        \caption{a \textbf{motorcycle} (\textit{4.80}) is parked on a dock with a bird perched on top of it}
    \end{subfigure}
    \hfill
    \begin{subfigure}[t]{0.2\textwidth}
        \centering
        \includegraphics[width=\textwidth]{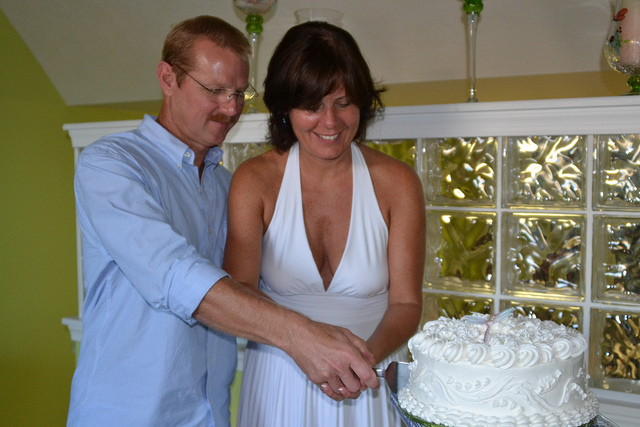}
        \caption{a bride and groom cutting their wedding \textbf{cake} (\textit{0.09})}
    \end{subfigure}
    \hfill
    \begin{subfigure}[t]{0.2\textwidth}
        \centering
        \includegraphics[width=\textwidth]{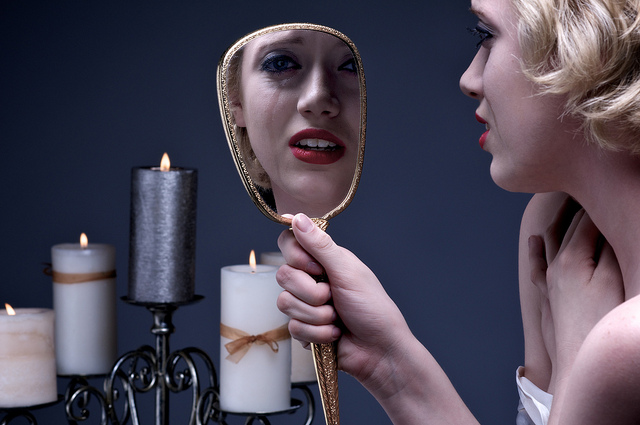}
        \caption{a woman holding a cup and a \textbf{cake} (\textit{5.29})}
    \end{subfigure}
    \vfill
    \begin{subfigure}[t]{0.2\textwidth}
        \centering
        \includegraphics[width=\textwidth]{figs/COCO_val2014_000000015017.jpg}
        \caption{a man standing on a tennis court \textbf{holding} (\textit{0.81}) a racquet}
    \end{subfigure}
    \hfill
    \begin{subfigure}[t]{0.2\textwidth}
        \centering
        \includegraphics[width=\textwidth]{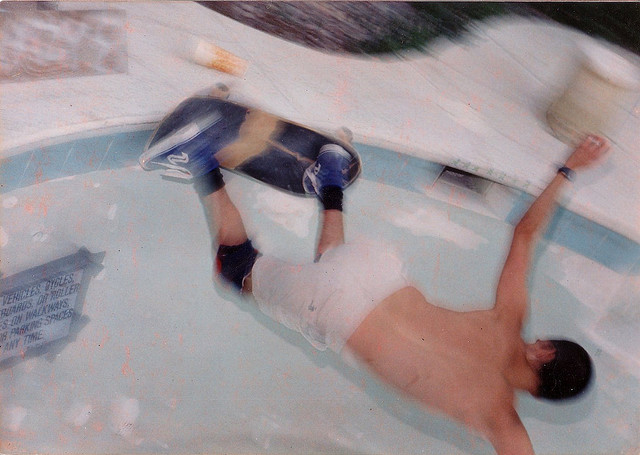}
        \caption{a young man is \textbf{holding} (\textit{4.76}) a skateboard in his hand}
    \end{subfigure}
    \hfill
    \begin{subfigure}[t]{0.2\textwidth}
        \centering
        \includegraphics[width=\textwidth]{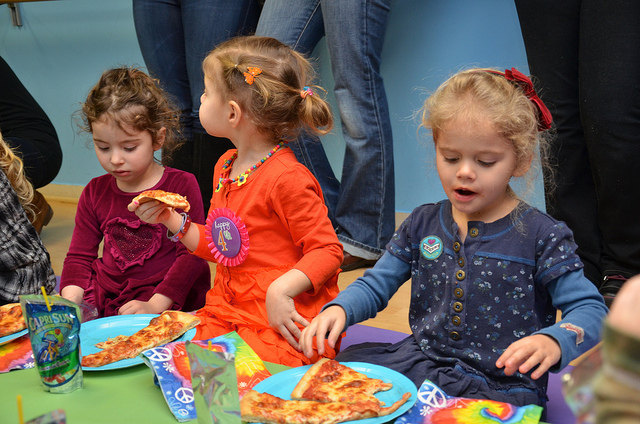}
        \caption{a group of children sitting at a table \textbf{eating} (\textit{1.00}) pizza}
    \end{subfigure}
    \hfill
    \begin{subfigure}[t]{0.2\textwidth}
        \centering
        \includegraphics[width=\textwidth]{figs/COCO_val2014_000000360730.jpg}
        \caption{a man is \textbf{eating} (\textit{4.01}) a hot dog at a restaurant}
    \end{subfigure}
    \caption{Examples of token predictions generated with the BUTD model with high and low uncertainty values for objects (top) and actions (bottom). Numbers in italic are predictive uncertainty values for the token predictions preceding them. The examples are cherry-picked.}
    \label{fig:examples}
\end{figure*}

\paragraph{Examples of predictions with high and low uncertainty}
Figure \ref{fig:examples} shows some example images and their captions generated from a BUTD model on the test set. The token predictions of interests and the corresponding uncertainty values are highlighted in bold and italic, respectively. We observe that highly uncertain predictions often correspond to unusual textures, features resembling the predicted tokens, or blurred images. For example, Figure \ref{fig:examples}(b) shows a motorcycle covered in vines; Figure \ref{fig:examples}(d) shows candles in the background which resemble cakes; Figure \ref{fig:examples}(f) is blurred.

\begin{table}[t]
    \centering
    \small
    \begin{tabular}{lcc}
    \toprule
         \multirow{2}{*}{\textbf{Model}} & \multicolumn{2}{c}{\textbf{Correlation coefficient}}  \\
         & epistemic & aleatoric \\
         \midrule
         FC & 0.313 & 0.299 \\
         BUTD & 0.334 & 0.228 \\
         Att2In & 0.360 & 0.268 \\
         Transformer & 0.269 & 0.131\\
         \bottomrule
    \end{tabular}
    \caption{Pearson correlation coefficients between hallucination and epistemic/aleatoric uncertainty in image captioning task. Epistemic uncertainty is more indicative of hallucination across four models.}
    \label{tab:corr}
\end{table}

\paragraph{Epistemic and aleatoric uncertainties}
As we could decompose the total uncertainty into two parts, we are interested in which part is more indicative of hallucination. Table \ref{tab:corr} shows the Pearson correlation coefficients between hallucination (binary) and epistemic/aleatoric uncertainty for all four models. We can see that both parts of uncertainty are weakly correlated with hallucination, while epistemic uncertainty is more indicative of hallucination across all four models compared to aleatoric uncertainty.

\section{Case Study: Data-to-text Generation}
Data-to-text generation \cite{kukich1983design,mckeown1992text} is a task to generate textual content conditioned on input content in the form of structured data such as tables. Neural models are prone to hallucination in data-to-text generation tasks compared to traditional template-based systems, and methods are proposed to improve faithfulness \cite{wiseman2017challenges,nie2019simple,tian2019sticking}. In this section, we discuss the relationship between predictive uncertainty and hallucination in data-to-text generation with ToTTo dataset \cite{parikh2020totto}.

\subsection{Generation Quality and Average Uncertainty}
We conduct token-level analysis in Section \ref{sec:img_cap}. Now we take a different route and analyze sentence-level quality with different average predictive uncertainty values. Experiment settings are described below.

\begin{table*}[t]
    \centering
    \small
    \begin{tabular}{lccccc}
    \toprule
         \textbf{Unc. Level} & \textbf{Avg Unc.} & \textbf{BLEU} & \textbf{Fluency (\%)} & \textbf{Faithfulness (\%)} & \textbf{Less/Neutral/More Coverage w.r.t. Ref}\\
         \midrule
         High & 1.83 - 3.74 & 10.2 & 46.0 & 41.3 & 79.4 / 15.9 / 04.7 \\
         Medium & 0.83 - 0.89 & 31.5 & 87.3 & 78.9 & 35.2 / 47.9 / 16.9 \\
         Low & 0.04 - 0.27 & 72.8 & 100.0 & 99.0 & 22.2 / 70.1 / 07.7\\
         \bottomrule
    \end{tabular}
    \caption{Evaluation results for candidates with high, medium, and low average predictive uncertainty values for ToTTo validation set. Unc. denotes uncertainty. Higher uncertainty candidates have lower quality and higher chance of being hallucinated/unfaithful w.r.t. the input tables.}
    \label{tab:totto}
\end{table*}

\paragraph{Dataset} ToTTo dataset consists of tables from English Wikipedia articles with their corresponding metadata, such as page title and section title. Candidate description texts are modified by annotators to pair with each table. Relevant table cells supporting the description texts are highlighted by the annotators as well. There are 120,761 table-text pairs in training, 7,700 in validation, and 7,700 in test. We use the baseline standard linearization approach to represent the highlighted portions of the tables along with their corresponding metadata (referred to as subtable with metadata in \cite{parikh2020totto}).

\paragraph{Model architecture and training} We use a standard sequence-to-sequence model with attention \cite{bahdanau2014neural,luo2018discriminability} for analysis. LSTM with 512 hidden size is used for both the encoder and the decoder. Adam optimizer with learning rate 1e-3 is used for the optimization. The model is trained with cross-entropy loss for 20 epochs. The checkpoint with the best validation loss is chosen for the evaluation. The implementation is done using fairseq \cite{ott2019fairseq}\footnote{https://github.com/pytorch/fairseq}.

\paragraph{Evaluation} We evaluate the average predictive uncertainty for all generated sentences in the validation set and select the top, bottom, and middle 5\% for comparison. BLEU score \cite{papineni2002bleu} is used as an automatic metric to evaluate the similarity to the references; further manual annotations are done to evaluate the \textit{fluency}, \textit{faithfulness} (precision), and \textit{coverage with respect to reference} (recall) of the generated sentences. Particularly, faithfulness reflects how likely the generated sentences hallucinate facts that are not supported by the tables. More details of the human evaluation metrics are described in \cite{parikh2020totto}. The goal is to measure how different the generation qualities are for candidates with varying average predictive uncertainties.

\subsection{Results and Discussions} Table \ref{tab:totto} summarizes the evaluation results for candidates with varying uncertainty values. It is obvious that candidates with higher average predictive uncertainty values are less fluent and more likely to contain hallucinations. Another interesting observation from Table \ref{tab:totto} is that the generated sentences with medium average uncertainty are more likely (16.9\%) to cover more table facts than the references compared to the ones with high (4.7\%) and low (7.7\%) average uncertainty. One possible explanation is that some table facts that are not always included in the references, when generated, have higher predictive uncertainty values than the facts that are almost always included in the references. Therefore, generated sentences with low uncertainty tend to include less but more confident facts considered by the model.

\section{Reducing Hallucination}

\subsection{Uncertainty-Aware Beam Search}
Because of the positive correlation between hallucination probability and predictive uncertainty, it is straightforward to incorporate uncertainty into the caption generation process to reduce hallucination. Beam search is the most used approximate decoding method in language generation. It keeps track of the top-$B$ scored candidates at each generation step and considers all single token extensions of the current candidates. 

More formally, denote the set of $B$ candidates in the beam at time step $t-1$ as $\mathcal{Y}_{t-1}=\{\vy^{(b)}_{t-1}\}_{b=1}^B$. All possible single token extensions of the candidates in $\mathcal{Y}_{t-1}$ form a set $\mathcal{C}_{t}=\{\vy\mid \vy_{t-1}\in\mathcal{Y}_{t-1}\land y_t\in\mathcal{V}\}$. Beam at step $t$ is then formed as:
\begin{align}
    \mathcal{Y}_{t}=&\argmax_{\vy_1\cdots\vy_B\in\mathcal{C}_t}\sum_{b=1}^B\log p(\vy_b|\vx)\nonumber\\
    &s.t.\quad \vy_i\neq\vy_j\quad\forall i\neq j
\end{align}

Uncertainty-aware beam search (UABS) adds a weighted penalty term in the beam search objective to balance between log probability and predictive uncertainty of the selected candidates. Let $u(\vy|\vx)$ be the function to measure the aggregated predictive uncertainty of candidate $\vy$ given input $\vx$, uncertainty-aware beam search updates the beam at step $t$ according to the following equation:
\begin{align}
    \mathcal{Y}_{t}=&\argmax_{\vy_1\cdots\vy_B\in\mathcal{C}_t}\sum_{b=1}^B\log p(\vy_b|\vx)-\lambda u(\vy_b|\vx)\nonumber\\
    &s.t.\quad \vy_i\neq\vy_j\quad\forall i\neq j
\end{align}
where $\lambda\geq 0$ is the weight controlling the degree to which we want to penalize decoding uncertainty. Larger $\lambda$ leads to candidates with smaller predictive uncertainty. In practice, this can be done by subtracting the weighted uncertainty term from the aggregated log probability scores at each decoding step before choosing top-$B$ candidates.

An important decision in using uncertainty-aware beam search is the choice of uncertainty term $u(\vy|\vx)$. We could use either the aleatoric or epistemic part of the predictive uncertainty or both. We compare these choices and discuss the results in the next section.

\begin{figure}[t]
    \centering
    \begin{subfigure}[t]{0.23\textwidth}
        \centering
        \includegraphics[width=\textwidth]{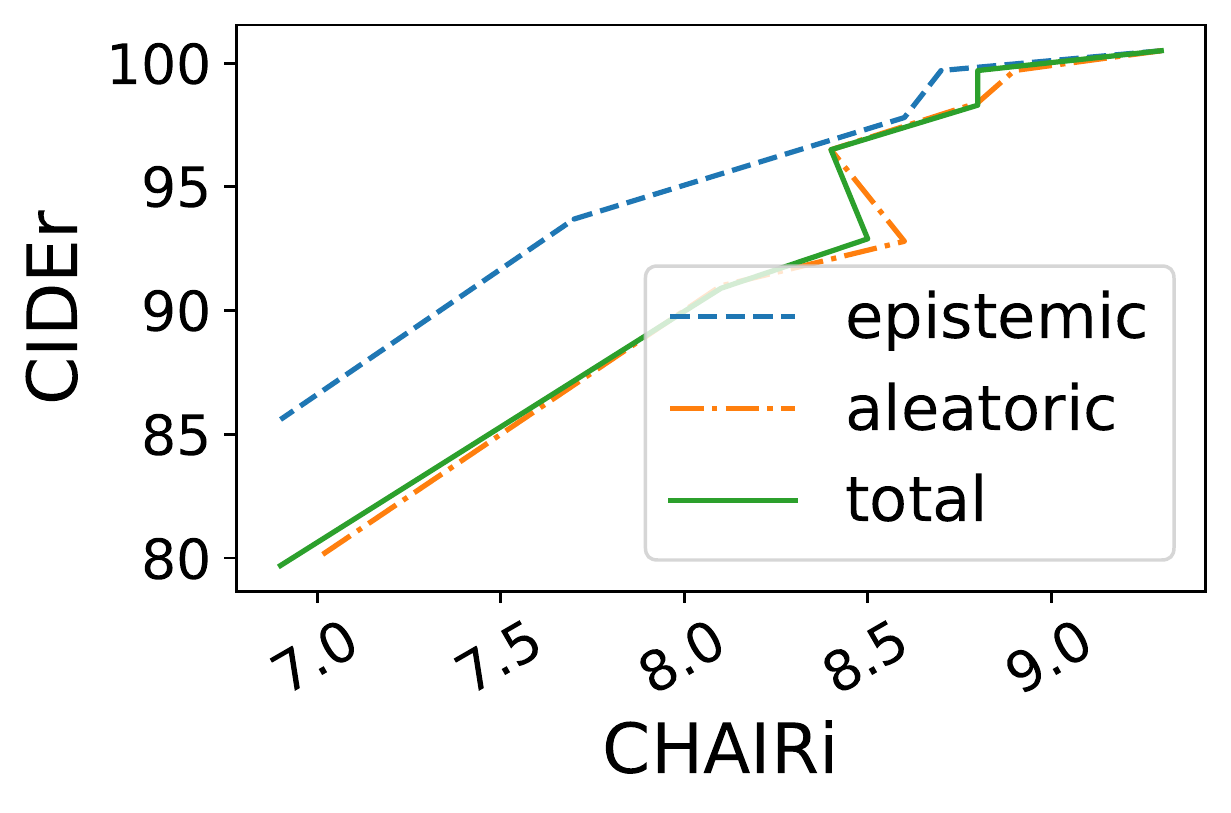}
        \caption{FC}
    \end{subfigure}
    \hfill
    \begin{subfigure}[t]{0.23\textwidth}
        \centering
        \includegraphics[width=\textwidth]{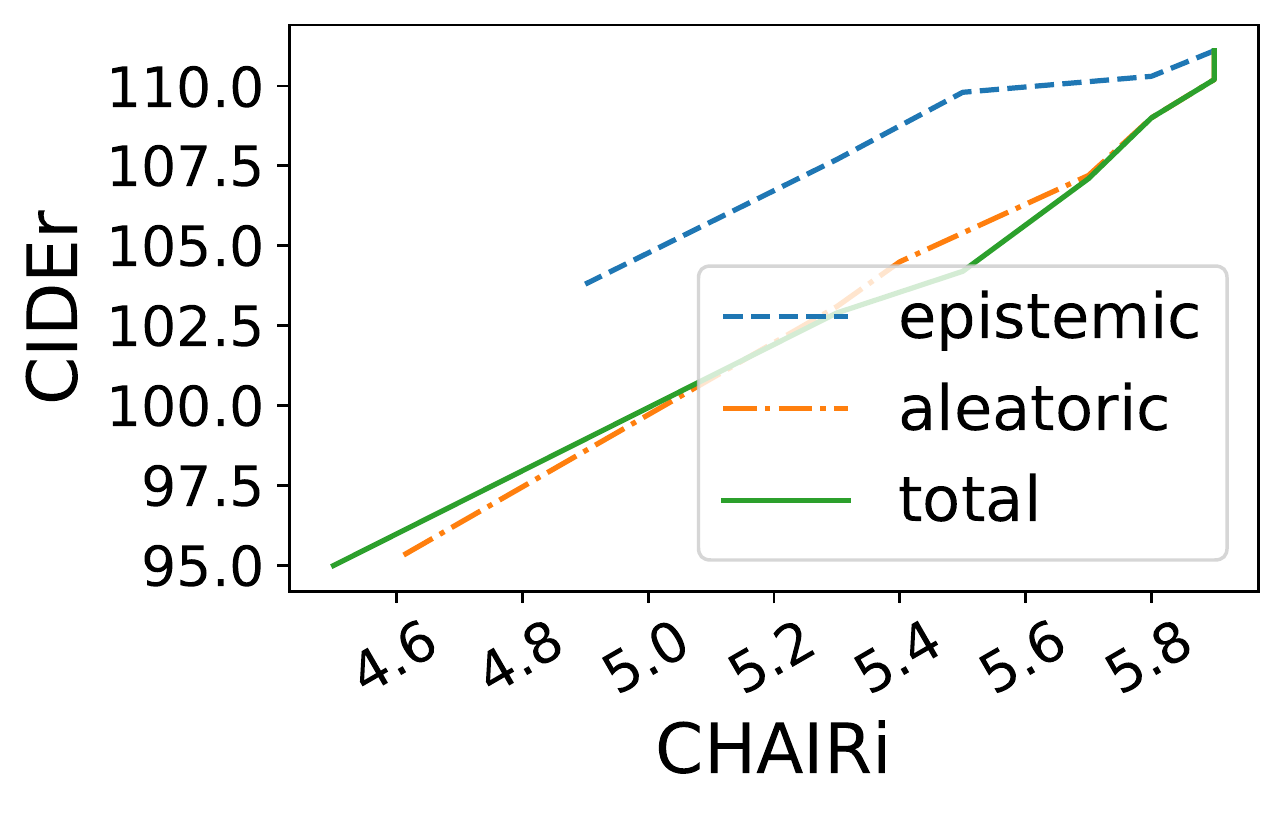}
        \caption{Att2In}
    \end{subfigure}
    \vfill
    \begin{subfigure}[t]{0.23\textwidth}
        \centering
        \includegraphics[width=\textwidth]{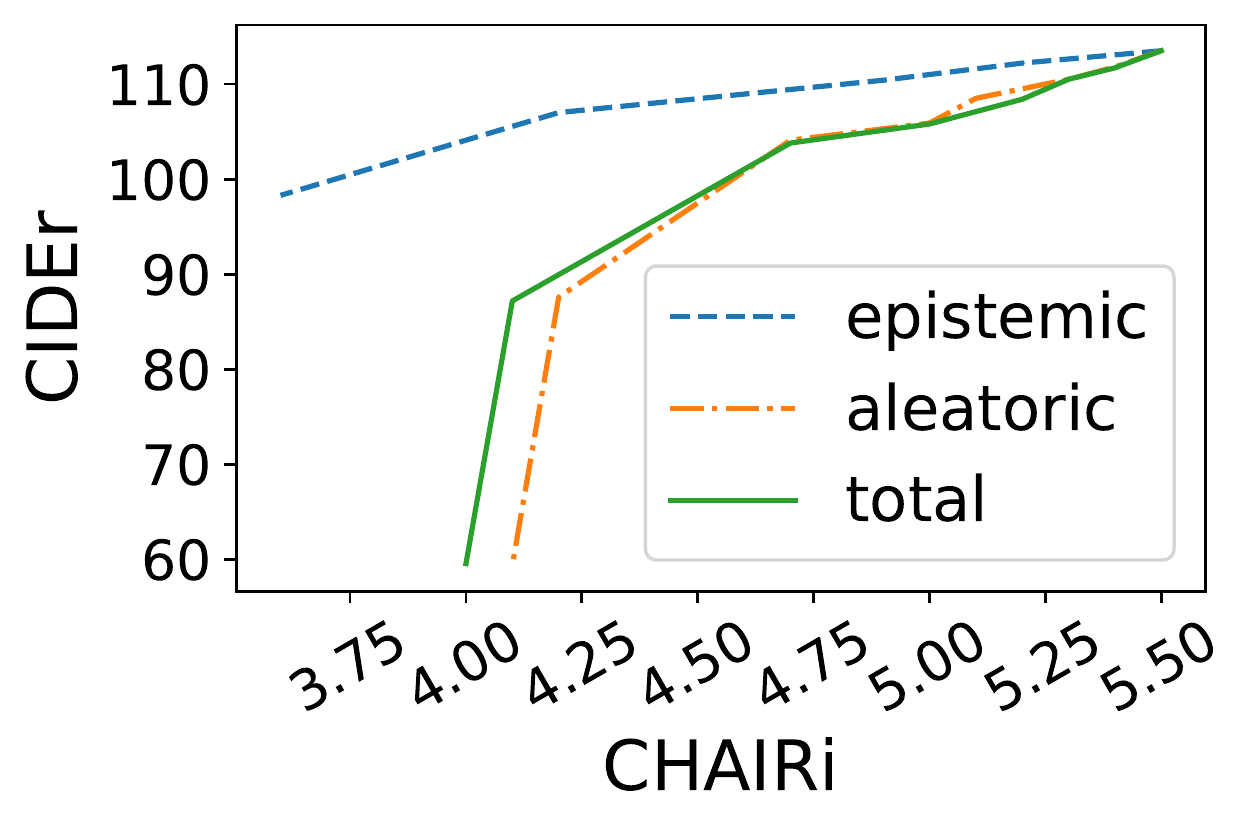}
        \caption{BUTD}
    \end{subfigure}
    \hfill
    \begin{subfigure}[t]{0.23\textwidth}
        \centering
        \includegraphics[width=\textwidth]{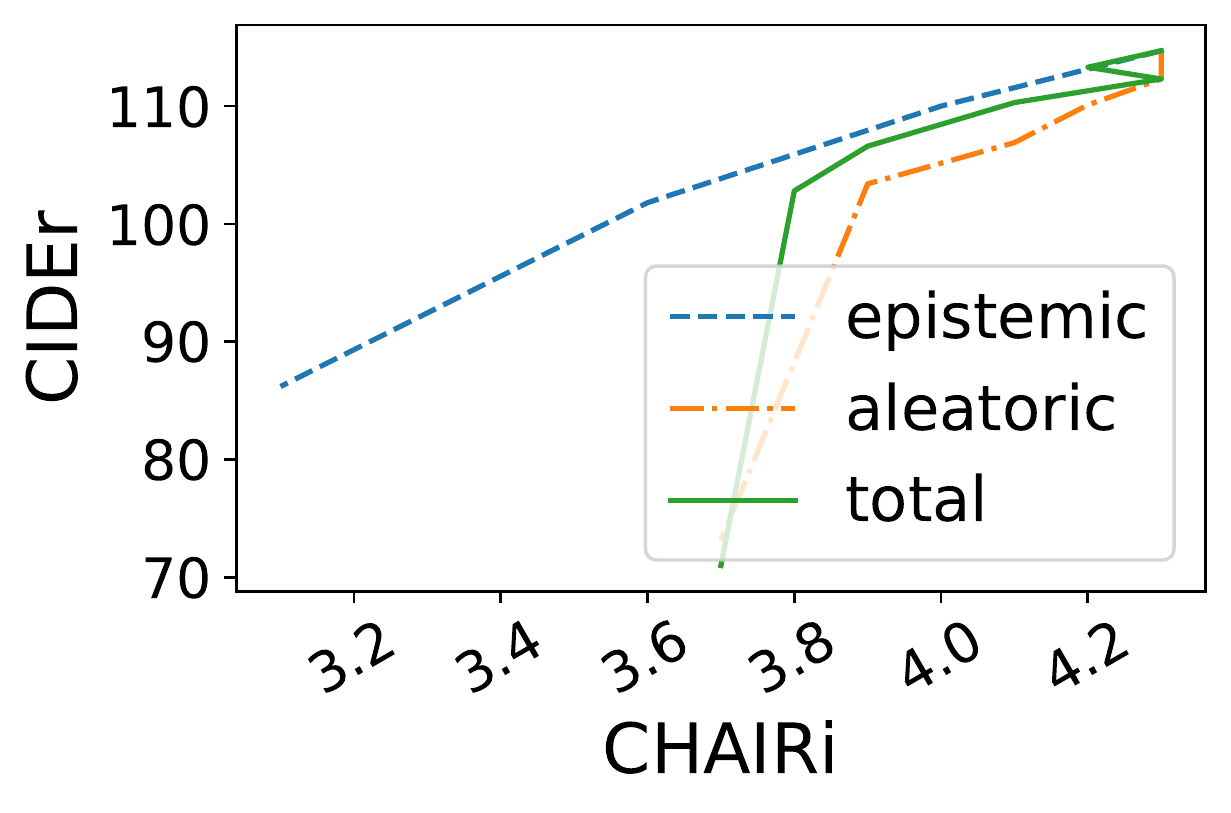}
        \caption{Transformer}
    \end{subfigure}
    \caption{CIDEr plotted against CHAIRi scores of captions generated with UABS with different uncertainty penalty weights. Lower CHAIRi score indicates less hallucination. Upper-left is better. Penalizing epistemic uncertainty in UABS achieves the best results.}
    \label{fig:tradeoffs}
\end{figure}

\begin{table*}[t]
    \centering
    \footnotesize
    \begin{tabular}{m{0.15\textwidth}m{0.22\textwidth}m{0.22\textwidth}m{0.22\textwidth}}
    \toprule
         \multirow{2}{*}{\textbf{Image}}  & \multicolumn{3}{c}{\textbf{UABS results with weight $\lambda$}}  \\
         & \multicolumn{1}{c}{0} & \multicolumn{1}{c}{20} & \multicolumn{1}{c}{80} \\
         \midrule
         \includegraphics[width=0.15\textwidth]{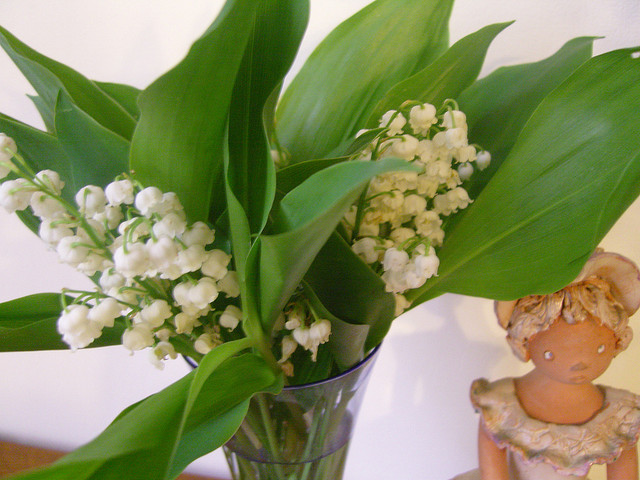} & \makecell[tl]{a vase filled with flowers\\ sitting on top of a table} & \makecell[tl]{a vase filled with lots of\\ white flowers}  & \makecell[tl]{there is a vase that has\\ flowers in it}  \\
         \midrule
         \includegraphics[width=0.15\textwidth]{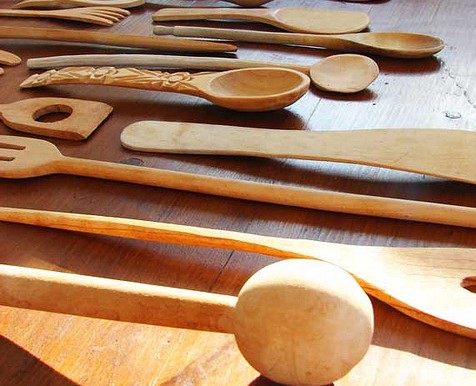} & \makecell[tl]{a wooden cutting board\\ topped with lots of food} & \makecell[tl]{a wooden cutting board\\ topped with lots of food}  & \makecell[tl]{a cutting board that has\\ a bunch on it}\\
         \bottomrule
    \end{tabular}
    \caption{Two examples of epistemic UABS results with varying penalty weights on the image captioning data set. In the first example the model successfully avoids hallucination of a table with $\lambda=20$ while in the second example it is unable to change the generated caption until larger penalty weight is set.}
    \label{tab:coco_ex}
\end{table*}

\subsection{Image Captioning Results}
With larger weights on the uncertainty penalty term, log probabilities of the decoded sentences drop. Therefore, we expect to see a trade-off between the quality of generated captions and the chance of hallucination. 

We empirically examine the trade-offs on the image captioning models with different uncertainty choices for the penalty term. 
We use a five-model ensemble for each of the four model architectures to estimate aleatoric and epistemic uncertainties. Due to the different magnitudes of aleatoric and epistemic uncertainties, we choose penalty weight $\lambda$ from $[0.1, 0.2, 0.4, 0.8, 1.0, 2.0, 4.0]$ for aleatoric and total uncertainty and $[10, 20, 40, 80]$ for epistemic uncertainty.

\begin{table}[t]
    \centering
    \small
    \begin{tabular}{lccccc}
    \toprule
         & $\lambda$ & \bf avg. len. & \bf \# obj. & \bf hal. \% & \bf gen. \%\\
         \midrule
         ref. & & 10.44 & 6114 & 0 & -\\
         
         base & 0 & 9.31 & 7328 & 5.5 & 0\\
         \midrule
         \multirow{4}{*}{epist.} & 10 & 9.21 & 7195 & 5.2 & 0\\
         & 20 & 9.16 & 7078 & 4.9 & 0.2\\
         & 40 & 9.15 & 6912 & 4.2 & 1.5\\
         & 80 & 9.12 & 6493 & 3.6 & 4.6\\
         \midrule
         \multirow{4}{*}{aleat.} & 0.1 & 9.32 & 7250 & 5.4 & 0\\
         & 0.4 & 9.32 & 7051 & 5.1 & 0 \\
         & 1.0 & 9.33 & 6800 & 4.7 & 1.0\\
         & 4.0 & 9.43 & 4349 & 4.1 & 28.4\\
         \bottomrule
    \end{tabular}
    \caption{Average sentence length and total number of objects detected in the captions generated by BUTD model with varying uncertainty penalty weight $\lambda$. Penalizing epistemic uncertainty leads to slightly shorter lengths. Number of objects mentioned by the captions decreases with increasing $\lambda$. \textbf{gen. \%} denotes percentage of generic responses. It is moderate with epistemic penalized results but can be very high if aleatoric uncertainty is heavily penalized.}
    \label{tab:length}
\end{table}

\begin{table*}[t]
    \centering
    \small
    \begin{tabular}{ccccc}
    \toprule
         \textbf{$\lambda$} & \textbf{BLEU} & \textbf{Fluency (\%)} & \textbf{Faithfulness (\%)} & \textbf{Less/Neutral/More Coverage w.r.t. Ref}\\
         \midrule
         0 & 40.1 & 92 & 79 & 34 / 60 / 6 \\
         10 & 33.6 & 83 & 84 & 41 / 51 / 8 \\
         20 & 27.4 & 73 & 80 & 52 / 42 / 6\\
         \bottomrule
    \end{tabular}
    \caption{Evaluation results for candidates decoded with different penalty weights for UABS on ToTTo validation set. Epistemic uncertainty is used for uncertainty penalization. Faithfulness first increases, then decreases to the same level as regular beam search results as we increase the penalty weight $\lambda$.}
    \label{tab:uabs_totto}
\end{table*}

Figure \ref{fig:tradeoffs} shows the trade-offs between CIDEr \cite{vedantam2015cider} and CHAIRi \cite{rohrbach2018object} scores of captions generated with uncertainty-aware beam search with different uncertainty choices and penalty weights. A smaller value of CHAIRi indicates the model is less likely to generate hallucinated objects, and a higher CIDEr indicates better caption quality. Therefore an approach that is to the upper left of another is better. As the penalty weight increases, we observe a decrease in both the CHAIRi and the CIDEr scores across all models. 

Table \ref{tab:coco_ex} shows two examples of different generated captions using epistemic UABS with varying penalty weights. In the first example, we can see that a medium penalty weight of 20 not only helps avoid the hallucination of a table but also adds correct information about the color of the flowers. In the second example, a medium penalty weight is unable to change the generated caption.

\begin{table*}[t]
    \centering
    \footnotesize
    \begin{tabular}{lccc}
    \toprule
         \multirow{2}{*}{\textbf{Reference}} & \multicolumn{3}{c}{\textbf{UABS results with weight $\lambda$}}  \\
         & 0 & 10 & 20 \\
         \midrule
         \makecell[tl]{barrows scored 164 net points\\ in virgin islands at the 2008\\ summer olympics.} &  \makecell[tl]{in virgin islands at the\\ 2008 summer olympics,\\ barrows \textcolor{blue}{iii} received 164\\ points.}& \makecell[tl]{in \textcolor{blue}{virgin islands} at the \\\textcolor{blue}{2008 summer olympics},\\ barrows received 164\\ points.} & \makecell[tl]{thomas barrows received\\ a total score of 164.}\\
         \midrule
         \makecell[tl]{janet gaynor won the first\\ academy award for best actress\\ for her performance in the\\ 7th heaven (1927 film).} & \makecell[tl]{janet gaynor won the\\ academy award for best\\ actress for \textcolor{red}{his} performance\\ in \textcolor{red}{janet gaynor}.} & \makecell[tl]{janet gaynor won the\\ academy award for best\\ actress.} & \makecell[tl]{janet gaynor won an \\academy award for best\\ actress.} \\
         \bottomrule
    \end{tabular}
    \caption{Two examples of UABS results with varying penalty weights on the ToTTo validation set. Blue tokens are correct table facts that are dropped by candidates generated with larger penalty weights; red tokens are incorrect/hallucinated facts that are dropped with larger penalty weights. In general, UABS with larger weights tend to produce sentences with less information that the model is more confident with.}
    \label{tab:totto_ex}
\end{table*}

Regarding the choice of uncertainty, it is notable that when penalizing epistemic uncertainty, the generated captions achieve higher CIDEr scores than penalizing aleatoric or total uncertainty. We hypothesize that epistemic uncertainty indicates the uncertainty of model weights. By penalizing epistemic uncertainty, we encourage the model to take the prediction path where it is well-calibrated. On the other hand, penalizing aleatoric uncertainty encourages the model to make low entropy predictions in all contexts regardless of the actual data distributions.

Table \ref{tab:length} shows the average sentence length, the number of objects, the percentage of hallucinations, and the percentage of generic responses in the captions generated by the BUTD model with different uncertainty choices and penalty weights on the test set. We can see that when penalizing epistemic uncertainty, UABS results in slightly shorter caption candidates. Both the number of objects and hallucination percentage decrease as we increase the weight $\lambda$. Interestingly, when penalizing aleatoric uncertainty, sentence length stays approximately the same despite lower CIDEr scores, as shown in Figure \ref{fig:tradeoffs}. Further investigation shows that this is partly due to an increasing number of generic captions such as ``\textit{there is no image here to provide a caption for}''. Penalizing epistemic uncertainty is much less likely to result in such generic captions. We can see that when increasing $\lambda$ from $1.0$ to $4.0$ with aleatoric UABS, the percentage of generic responses jumps drastically from $1.0\%$ to $28.4\%$. In comparison, epistemic UABS keeps the generic response rates low while achieving lower hallucination rates.

\subsection{Data-to-text Results}
We also evaluate the effect of UABS on the ToTTo dataset. We choose to penalize epistemic uncertainty due to its better performances than aleatoric uncertainty, as shown in the previous section. A five-model deep ensemble is used to quantify the epistemic uncertainty and generate results with UABS. We compare the BLEU score and three human evaluation metrics among results generated with different uncertainty penalty weights. 100 generation results are randomly selected and evaluated for each penalty weight choice. The results are shown in Table \ref{tab:uabs_totto}. We can see that a relatively small penalty weight leads to a reduced hallucination chance (hence more faithful) with a cost on the BLEU score and fluency.

To qualitatively examine the sentences generated with different $\lambda$ values, we show example results on the ToTTo validation set in Table \ref{tab:totto_ex}. 
We can see that with larger penalty weights, the UABS results drop certain statements that the model deems less confident regardless of the correctness. This results in shorter but more confident predictions for UABS results with a larger uncertainty penalty.

\section{Related Work}
\paragraph{Hallucination} 
There are many pieces of anecdotal evidence of hallucination presented in various NLG tasks. Most recently, researchers started investigating the phenomenon systematically.
\citet{rohrbach2018object} analyzes object hallucination focusing on the objects that appeared in the MSCOCO segmentation challenge. They propose the CHAIR metric to quantify the severity of object hallucination. They find that the models tend to make predictions consistent with a language model trained on the captions instead of a model trained to predict objects in an image. Therefore hallucination is caused by an over-reliance on the language priors.
\citet{nie2019simple} believes that the origin of the hallucination problem in neural surface realization comes from the data side. More specifically,
datasets used for NLG systems often include instances with information misalignment between the input structure and the output text. They propose integrating a language understanding module for iterative data refinement to better align meaning representations and output text. 
\citet{muller2019domain} examines hallucination in neural machine translation and observes that the phenomenon is most common in out-of-domain settings. They empirically compare several strategies to improve domain robustness in NMT and find that a combination of reconstruction and a noisy channel model for reranking is most effective. 

These observations are consistent with our findings. For example, domain shift and data misalignment are known to lead to a higher level of epistemic uncertainty \cite{kendall2017uncertainties} which makes hallucination a more severe problem.
\paragraph{Uncertainty quantification}
Uncertainty quantification has attracted more attention recently due to the progress in Bayesian deep learning. Bayes by backprop \cite{blundell2015weight}, Monte Carlo dropout \cite{gal2016dropout}, and deep ensembles \cite{lakshminarayanan2017simple} are examples of popular Bayesian approaches to evaluate uncertainty with deep neural models. 
\citet{kendall2017uncertainties} investigates the benefits of modeling epistemic and aleatoric uncertainty in vision tasks such as semantic segmentation and depth regression. They show that it is important to model aleatoric uncertainty with large datasets and real-time applications and epistemic uncertainty with small datasets and safety-critical applications. Other applications of uncertainty quantification have been explored in the context of time series predictions \cite{zhu2017deep}, natural language processing tasks \cite{xiao2019quantifying}, etc. More broadly, prediction entropy has been analyzed in different neural language generation tasks \cite{ott2018analyzing,xu2020understanding}.
\citet{depeweg2018decomposition} shows how to extract and decompose uncertainty in Bayesian neural networks with latent variables for decision-making purposes. They show that active learning and risk-sensitive reinforcement learning both benefit from uncertainty decomposition.

\section{Discussion and Conclusions}

We investigate the relationship between hallucination and predictive uncertainty in image captioning and data-to-text generation tasks and show that predictions with higher uncertainty are more prone to hallucination. In particular, epistemic uncertainty is more indicative of hallucination than aleatoric uncertainty.
We propose uncertainty-aware beam search to incorporate uncertainty into the decoding process to reduce hallucination. We show that uncertainty decomposition helps the proposed beam search variant to achieve a better performance-hallucination trade-off. Specifically, penalizing epistemic uncertainty yields better results compared to penalizing aleatoric or total uncertainty.

In this work, we analyze uncertainty from the token level. This might be restrictive because uncertainty corresponds to the current prediction context instead of the predicted token. The relationship between hallucination and uncertainty, therefore, can be much more complicated than a linear one. It is still possible to produce hallucinated information with a very confident model. The proposed UABS reduces hallucination by limiting the total uncertainty of the generated text. As a result, it might lead to shorter generations and lower generation quality. Devising more sophisticated uncertainty-aware training and decoding methods with less adverse effects on the generation quality is a future direction to explore.

\section*{Acknowledgement}

This work was supported by the National Science Foundation award \#2048122. The views expressed are those of the author and do not reflect the official policy or position of the US government. 
\bibliographystyle{acl_natbib}
\bibliography{eacl2021}

\end{document}